\documentclass[letterpaper, 10 pt, conference]{ieeeconf}  

\IEEEoverridecommandlockouts                              
\usepackage{graphicx} 
\usepackage{epsfig} 
\usepackage{times} 
\usepackage{amsmath} 
\usepackage{amssymb}  
\usepackage{cite}
\usepackage{bm}
\usepackage{booktabs}
\usepackage{cases}
\usepackage{xcolor}
\usepackage{flushend}
\usepackage{algorithm}
\usepackage{algpseudocode}
\usepackage{subfigure}
\makeatletter

\newtheorem{remark}{Remark}

\newcommand{\Rmnum}[1]{\expandafter\@slowromancap\romannumeral #1@}
\makeatother
\def\BibTeX{{\rm B\kern-.05em{\sc i\kern-.025em b}\kern-.08em
    T\kern-.1667em\lower.7ex\hbox{E}\kern-.125emX}}
\usepackage{stfloats}

\title{\LARGE \bf
Underwater Motions Analysis and Control of a Coupling-Tiltable Unmanned Aerial-Aquatic Vehicle
}

\author{Dongyue Huang$^{1,2}$, Minghao Dou$^{2}$, Xuchen Liu$^{2}$, Tao Sun$^{1}$, Jianguo Zhang$^{1}$, Ning Ding$^{1}$, \\Xinlei Chen$^{3}$ and Ben M. Chen$^{2}$
\thanks{This work was supported in part by the Grant 2024A0505040007, and in part by the InnoHK of the Government of the Hong Kong Special Administrative Region via the Hong Kong Centre for Logistics Robotics.}
\thanks{$^{1}$Shenzhen Institute of Artificial Intelligence and Robotics for Society (AIRS), Shenzhen, China. {\tt\small \{huangdongyue, suntao, zhangjianguo, dingning\}@cuhk.edu.cn}}%
\thanks{$^{2}$Department of Mechanical and Automation Engineering, The Chinese University of Hong Kong, Shatin, N.T., Hong Kong. {\tt\small \{dyhuang,}
{\tt\small mhdou,xcliu,bmchen\}@mae.cuhk.edu.hk}}%
\thanks{$^{3}$Shenzhen International Graduate School, Tsinghua University, Shenzhen, China. {\tt\small chen.xinlei@sz.tsinghua.edu.cn}}%
}

\begin{document}

\maketitle
\thispagestyle{empty}
\pagestyle{empty}

\begin{abstract}

Coupling-Tiltable Unmanned Aerial-Aquatic Vehicles (UAAVs) have gained increasing importance, yet lack comprehensive analysis and suitable controllers. This paper analyzes the underwater motion characteristics of a self-designed UAAV, Mirs-Alioth, and designs a controller for it. The effectiveness of the controller is validated through experiments. The singularities of Mirs-Alioth are derived as Singular Thrust Tilt Angle (STTA), which serve as an essential tool for an analysis of its underwater motion characteristics.
The analysis reveals several key factors for designing the controller. These include the need for logic switching, using a Nussbaum function to compensate control direction uncertainty in the auxiliary channel, and employing an auxiliary controller to mitigate coupling effects. Based on these key points, a control scheme is designed. It consists of a controller that regulates the thrust tilt angle to the singular value, an auxiliary controller incorporating a Saturated Nussbaum function, and a logic switch. Eventually, two sets of experiments are conducted to validate the effectiveness of the controller and demonstrate the necessity of the Nussbaum function.

\end{abstract}

\section{Introduction}
With advancements in sensor technology and manufacturing techniques, robotics has been embraced in a multitude of fields with the aim of reducing the demands on human labor. The integration of aerial and underwater robotics holds the potential to increase the efficiency and efficacy of various tasks, particularly in the realm of underwater search and rescue operations \cite{Tan2021US}. Many innovative unmanned aerial-aquatic vehicles (UAAVs) have been designed, such as those in \cite{Maia, liu2023mirsx, bi2024, liu2023tj}.

\begin{figure}[t!]
        \centering
		\subfigure[\label{fig:front1}] 
{\includegraphics[width = 0.205\textwidth]{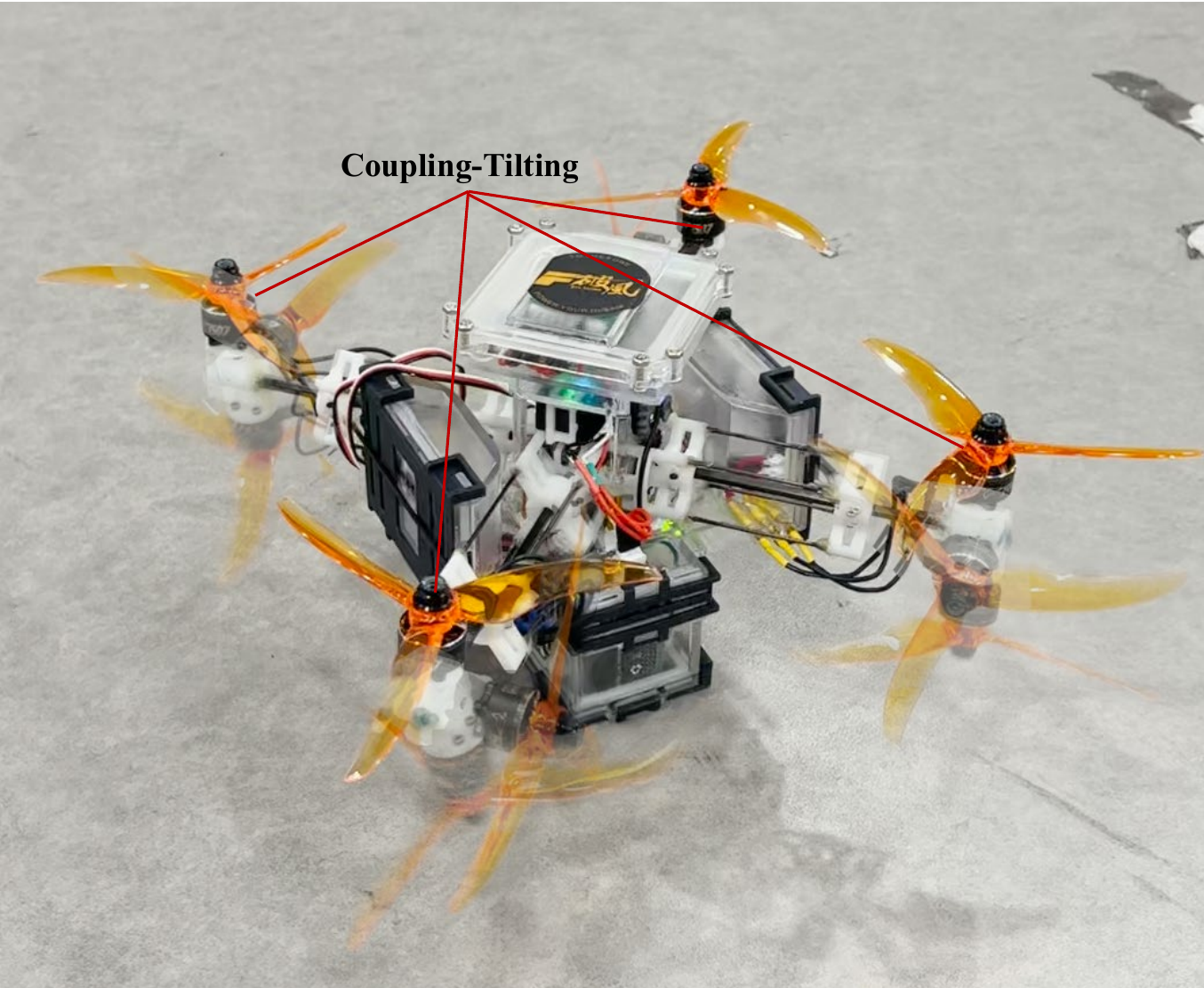}}~
		\subfigure[\label{fig:front2}]
{\includegraphics[width = 0.26\textwidth]{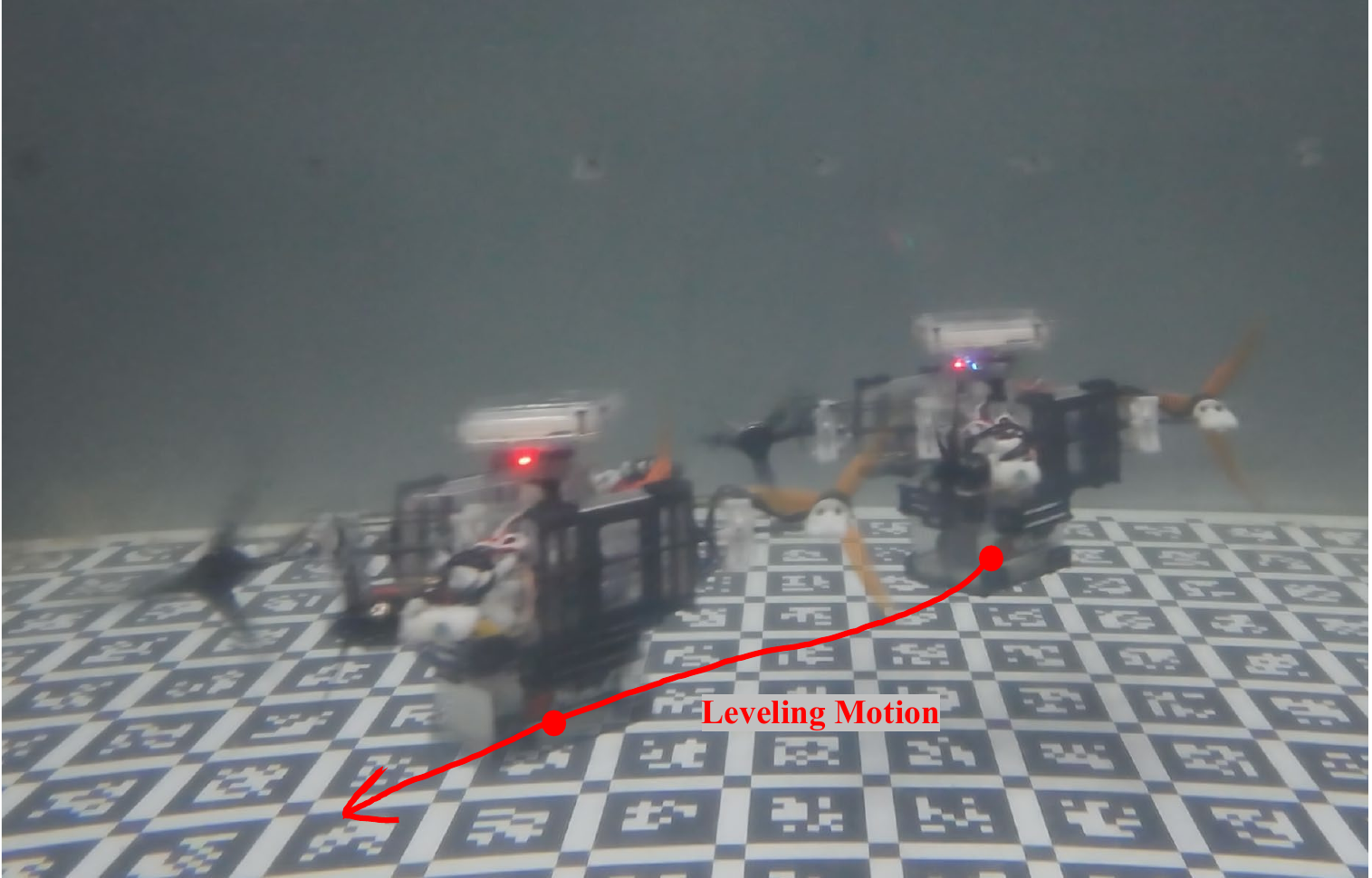}}
\caption{Mirs-Alioth. (a) Tilting mechanism diagram. (b) Snapshot of the experiment when operating a leveling motion under the proposed controller. }
\label{fig:front}
\end{figure}

Considering both improved maneuverability for underwater locomotion and a proper weight for aerial flight, a novel UAAV called ``Mirs-Alioth'', is designed based on our proposed design method \cite{huang2024sys}. As it has a tiltable component to change the all rotors plate's direction in a coupling mechanically-linked manner while keeping the whole body static \cite{tan2}, as shown in Fig. \ref{fig:front1}. Therefore, it is classified as a coupling-tiltable UAAV.
Our previous published work \cite{tan3} has mentioned and roughly discussed that the vehicle can generate leveling motions, like surge motion and spiral motion with invariant pitch or roll angle. These motions are unprocurable to the fixed-tilted UAAV but crucial. Because large attitude changes can disrupt onboard sensors like cameras, adversely affecting localization algorithm. Additionally, significant attitude changes can result in the loss of tracking or inspection targets, complicating exploration and monitoring tasks. However, unfortunately, previous work on Mirs-Alioth has not provided a comprehensive analysis that clearly defines control objectives based on motion characteristics to support controller design. 
Additionally, due to the unique tilting mechanism of coupling-tiltable UAAVs, there has been little contribution from other researchers in the area of motion analysis for this type of vehicle.

Many researchers have conducted extensive work on the control of tiltable vehicles, including UAAVs, ROVs, and UAVs.
The control research on tiltable UAVs has developed over many years, resulting in numerous mature studies, such as \cite{allenspach2020design}. However, to the best of our knowledge, there has been little work focused on controlling coupling-tiltable UAVs. Most research in this area has concentrated on systems where each rotor tilts independently or in pairs, but not on systems where all rotors tilt coupling.
For tiltable ROVs, the only work that has addressed control for coupling-tiltable ROVs, to our knowledge, is \cite{jin2015six}. Unfortunately, their approach uses fixed control angles for the tilting surfaces, rather than continuous control. This method can lead to undesirable consequences, such as system oscillation or even divergence.
A similar issue is seen in the coupling-tiltable UAAV. Their control strategy also adopts a discrete tilting mechanism, with other aspects controlled by conventional PID methods, as used in typical quadrotor  systems (see \cite{rao2023puffin,li2023aerial}).
It is evident from the current research on coupling-tiltable vehicles that there is still a significant gap in this area of control development.

To this end,
this paper builds upon the work in \cite{tan3} by conducting a more comprehensive and detailed analysis of the underwater motion characteristics of Mirs-Alioth based on its unique model. These characteristics and constraints are categorized as Singular Thrust Tilt Angle (STTA).
Based on the motion analysis, a controller leveraging singularities is designed, accompanied by a logic switching algorithm to toggle between different channels. To address the coupling effects that arise during complex motions between different channels, an auxiliary controller is introduced. Additionally, to mitigate the control direction uncertainty caused by the tilting mechanism, a Saturated Nussbaum function is incorporated into the auxiliary controller for stabilization. In this paper, the vehicle operates under the proposed controller is able to conduct leveling motion with continuous thrust tilt angle control. The effectiveness of the designed controller is validated through a series of experiments, as shown in Fig.~\ref{fig:front2}.

The remainder of the article is organized as follows. 
Section~\ref{background} and Section~\ref{modeling} illustrate the mathematical model. Section~\ref{analysis} introduces the derivation of STTAs and analysis of underwater motion based on STTAs. Section~\ref{controller} introduces the details about the proposed controller. Lastly, experiment validations are conducted in Section~\ref{validation}.

\section{Background}\label{background}
In this section, some predefined parameters are demonstrated as background knowledge used for the latter analysis, followed the benchmark proposed by Fossen \cite{b14}. The inertial matrix $\bm{M}$ is organized as:

\begin{footnotesize}
\begin{equation}
    \bm{M} = \left[ {\begin{array}{*{20}{c}}
    {{M_{11}}}&0&0&0&{{M_{51}}}&0\\
    0&{{M_{22}}}&0&{{M_{41}}}&0&0\\
    0&0&{{M_{33}}}&0&0&0\\
    0&{{M_{14}}}&0&{{M_{44}}}&0&0\\
    {{M_{15}}}&0&0&0&{{M_{55}}}&0\\
    0&0&0&0&0&{{M_{66}}}
    \end{array}} \right],
    \label{eq_M}
\end{equation}
\end{footnotesize}%
where ${M_{11}} = m - {X_{\rm{\dot u}}}$, ${M_{22}} = m - {Y_{\rm{\dot v}}}$, ${M_{33}} = m - {Z_{\rm{\dot w}}}$, ${M_{44}} = I_{\rm{xx}} - {K_{\rm{\dot p}}}$; ${M_{55}} = I_{\rm{yy}} - {M_{\rm{\dot q}}}$; ${M_{66}} = I_{\rm{zz}}- {N_{\rm{\dot r}}}$, ${M_{15}} = {M_{51}} = mz_{\rm{g}} - {X_{\rm{\dot q}}}$, ${M_{14}} = {M_{41}} = -mz_{\rm{g}} - {Y_{\rm{\dot p}}}$, $m$ denotes the mass of the vehicle, $z_{\rm{g}}$ denotes the distance between center of gravity (CoG) and center of buoyancy (CoB), ${\bm{I}} = {\rm diag}([{I_{{\rm{xx}}}}{\rm{, }}{I_{{\rm{yy}}}}{\rm{, }}{I_{{\rm{zz}}}}])$ is the moment of inertia of the vehicle.
The damping term $\bm{D}$ neglecting the nonlinear damping is defined as $\bm{D} =  {\rm diag}\{ {X_{\rm{u}}},{Y_{\rm{v}}},{Z_{\rm{w}}},{K_{\rm{p}}},{M_{\rm{q}}},{N_{\rm{r}}}\}$.
    
The Coriolis/centripetal matrix $\bm{C}(\bm{\nu})$ consisting of the rigid body part and the hydrodynamic part is defined as:

\begin{footnotesize}
\begin{equation}
    \bm{C}(\bm{\nu})= \left[ {\begin{array}{*{20}{c}}
    0&0&0&{{C_{11}}}&{{C_{12}}}&{{C_{13}}}\\
    0&0&0&{{C_{14}}}&{{C_{15}}}&{{C_{16}}}\\
    0&0&0&{{C_{17}}}&{{C_{18}}}&0\\
    { - {C_{11}}}&{{C_{12}}}&{ - {C_{17}}}&0&{{C_{21}}}&{{C_{22}}}\\
    {{C_{14}}}&{ - {C_{15}}}&{ - {C_{18}}}&{{C_{23}}}&0&{{C_{24}}}\\
    { - {C_{13}}}&{ - {C_{16}}}&0&{{C_{25}}}&{{C_{26}}}&0
    \end{array}} \right],
    \label{eq_C}
\end{equation}
\end{footnotesize}%
where $C_{11} = C_{15} = mz_{\rm{g}}r$, $C_{12} = -C_{14} = mw - {Z_{\rm{{\dot w}}}}  $, $C_{13} = -mv + {Y_{\rm{{\dot v}}}}v + {Y_{\rm{{\dot p}}}}p $, $C_{16} = mu - {X_{\rm{{\dot u}}}}u - {X_{\rm{{\dot q}}}}q$, $C_{17} = -m(z_{\rm{g}}p-v) - {Y_{\rm{{\dot v}}}}v - {Y_{\rm{{\dot p}}}}p$, $C_{18} = -m(z_{\rm{g}}p+u) + {X_{\rm{{\dot u}}}}u + {X_{\rm{{\dot q}}}}q$, $C_{21} = -C_{23} = {I_{{\rm{zz}}}}r - {N_{\rm{{\dot r}}}}r$, $C_{22} = -C_{25} = -{I_{{\rm{yy}}}}q + {M_{\rm{{\dot q}}}}q$, $C_{24} = -C_{26} = {I_{{\rm{xx}}}}p - {K_{\rm{{\dot p}}}}p$.

The generalized restoring force matrix $\bm{g} ({\mit{\bm{\eta}}})$, vectored by the gravitational and buoyancy force is organized as $\bm{g} = [ { (G - B)s_\theta },{-(G - B)c_\theta s_\phi },{-(G - B)c_\theta c_\phi }, { {z_{\rm{g}}}G \cdot c_\theta s_\phi },\\{ {z_{\rm{g}}}G \cdot s_\theta },0]^{\rm{T}}$, where $s_\theta$ denotes $\rm{sin}(\theta)$, $c_\theta$ denotes $\rm{cos}(\theta)$, $t_\theta$ denotes $\rm{tan}(\theta)$, so as $\phi$ and $\psi$. $G$ and $B$ denote the gravity and buoyancy of the vehicle respectively.

\section{Modelling}\label{modeling}
In this section, we further introduce the nonlinear model and the decoupled model of our vehicle, based on the background knowledge. Tow assumptions are given following for the analysis later. Mirs-Alioth, as a morphable robot, exhibits different dynamics compared to unmorphable robots, but these variations can be treated as disturbance. Then first assumption is made. Additionally, in the subsequent discussion of this paper, motions without significant attitude changes will be considered. Therefore the second one is given.

\begin{enumerate}
    \item CoG/CoB and moment of inertia are time-invariant.
    \item The approximations are made as \( s_\theta \approx \theta \), \( s_\phi \approx \phi \), \( c_\theta \approx 1 \), and \( c_\phi \approx 1 \).
\end{enumerate}

\subsection{Nonlinear Mathematical Model}
\begin{figure}[h]
    \centering
    \includegraphics[width=0.7\linewidth]{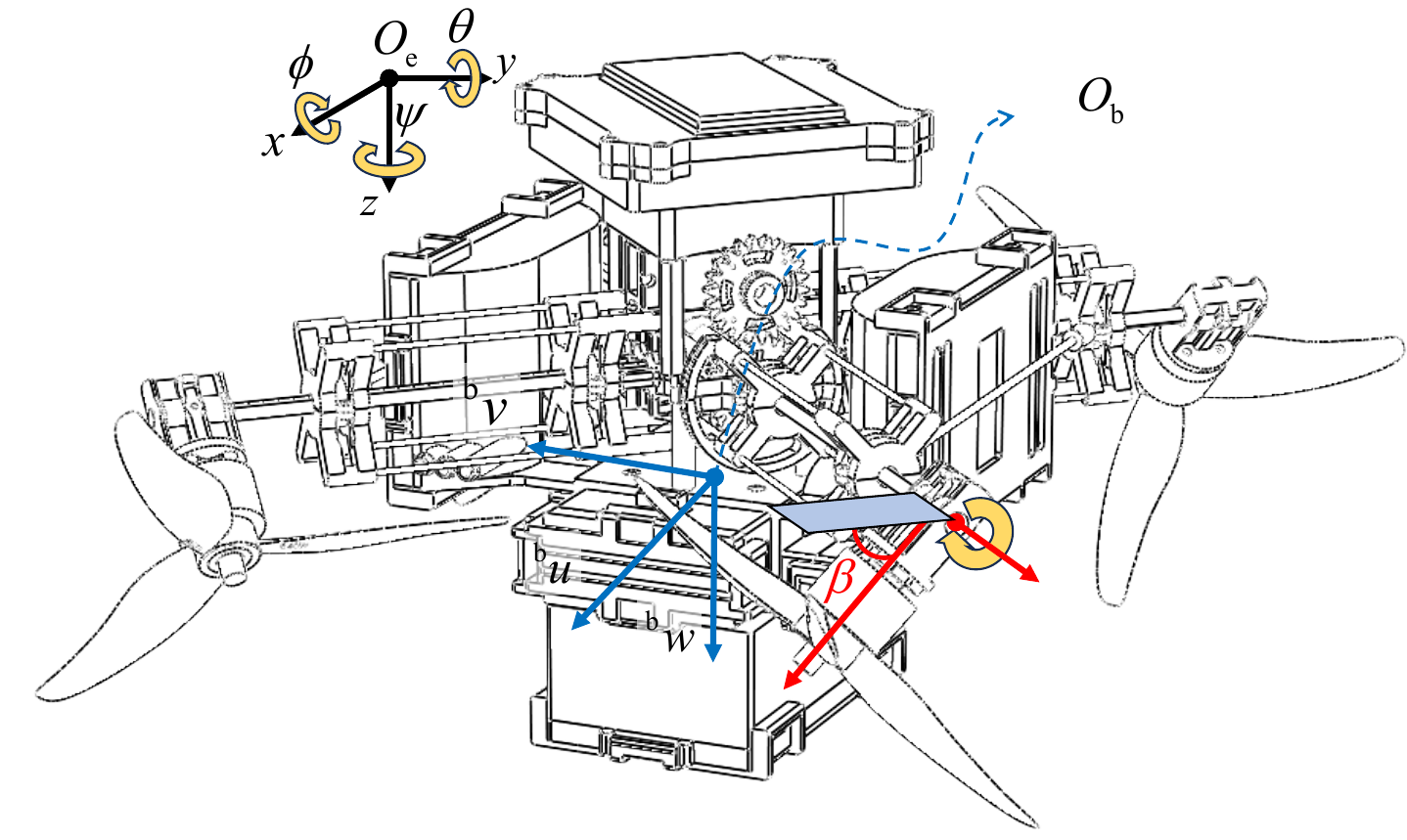}
    \caption{Coordinates of Mirs-Alioth}
    \label{fig:frame}
\end{figure}

Assuming the vehicle is fully submerged, Fig.~\ref{fig:frame} introduces two reference frames: the body reference frame (b-frame) affixed to the vehicle body with its origin at the CoB, denoted as \(O_{\text{b}}\), and the inertial reference frame (e-frame) for tracking global motion, with its origin at \(O_{\text{e}}\).

As in \cite{huang2024first}, $\bm{\nu} = [{^{\rm{b}} \bm{v}} \ {^{\rm{b}} \bm{\omega}}]^{\rm{T}} $ as the velocity, represented in the b-frame is defined.  
The body-fixed linear velocity is denoted as ${^{\rm{b}} \bm{v}} = [u\ v\  w]^{\rm{T}}$.
And ${^{\rm{b}} \bm{\omega}} = [p\ q\ r]^{\rm{T}}$ denotes the body-fixed angular velocity, the pose variable ${\mit{\bm{\eta}}} = [\bm{P}_{\rm{e}}\ {\bm{\Theta}}_{\rm{e}}]^{\rm{T}}$, consisted by the position $\bm{P}_{\rm{e}} = [x\ y\ z]^{\rm{T}}$ and the attitude with Euler angles labeled as ${\bm{\Theta}}_{\rm{e}} = [\phi\ \theta\ \psi]^{\rm{T}}$, both represented in e-frame. We define
$\bm{J} = diag ([{\bm{R}}_{\rm{b}}^{\rm{e}},\bm{S}^{{\rm{ - 1}}}])$ is the transformation matrix, consisting of $\bm{R}_{\rm{b}}^{\rm{e}}$, the rotation matrix from the body frame to the world frame and $\bm{S}^{{\rm{ - 1}}}$, the lumped transformation matrix of angular velocity in \cite{b10}.

Then the nonlinear model can be described as,
\begin{equation}
    \dot {\mit{\bm{\eta}}}  = \bm{J}\bm{\nu},
    \label{eq2}
\end{equation}
\begin{equation}
    \bm{M}\dot{\bm{\nu}} + \bm{C}(\bm{\nu})\bm{\nu} + \bm{D}\bm{\nu} + \bm{g} ({\bm{\eta}}) = {\bm{\tau}}.
    \label{eq1}
\end{equation}

Besides, ${\mit{\bm{\tau}}} = [\bm{F}_{\rm{b}}\ \bm{M}_{\rm{b}}]^{\rm{T}} $ is the vector of generalized force and moment, following by ${\bm{\tau}} = \bm{B}(\beta) \bm{u}$. Here, $\beta$ denotes the symmetrically thrust tilt angle ranging from $-{\pi}/2$ to ${\pi}/2$ \cite{tan3}, represented in the b-frame, as shown in Fig. \ref{fig:frame}.
We further define the input $\bm{u} = {\varpi _i^2}(i \in \{1,2,3,4\})$, with ${\varpi _i}$ denoting the rotational speed of the motor. Then the Control Effectiveness Matrix (CEM) is reorganized as
\begin{equation}
\begin{aligned}
    \bm{B}(\beta) = \frac{\sqrt 2}{2} \left[ {\begin{array}{*{20}{c}}
    {c_\beta {K_{\rm{T}}} \cdot {e_{{\rm{x}}}}}\\
    {c_\beta {K_{\rm{T}}} \cdot {e_{{\rm{y}}}}}\\
    {-\sqrt 2 s_\beta {K_{\rm{T}}} \cdot {e_{{\rm{z}}}}}\\
    {{{k_1}(\beta )} \cdot {e_{{\rm{\phi}}}}}\\
    {{{k_2}(\beta )} \cdot {e_{{\rm{\theta}}}}}\\
    {{\sqrt 2 {k_3}(\beta )} \cdot {e_{{\rm{\psi}}}}} 
    \end{array}} \right],
    \label{eq7}
\end{aligned}
\end{equation}
with
\begin{equation}
  {k_1}(\beta ) = l \cdot s_\beta  \cdot {K_{\rm{T}}} + c_\beta ({K_{\rm{M}}} - {z_{\rm{t}}}{K_{\rm{T}}}),
  \label{eq_B1}
\end{equation}
\vspace{-12pt}
\begin{equation}
  {k_2}(\beta ) =  - l \cdot s_\beta  \cdot {K_{\rm{T}}} + c_\beta ({K_{\rm{M}}} + {z_{\rm{t}}}{K_{\rm{T}}}),  
  \label{eq_B2}
\end{equation}
\vspace{-12pt}
\begin{equation}
  {k_3}(\beta ) = {K_{\rm{T}}}l \cdot c_\beta  - s_\beta  \cdot {K_{\rm{M}}}, 
  \label{eq_B3}
\end{equation}
where $k_i(\beta)$ is the coefficient of the body moment, related to the thrust tilt angle $\beta$, wheelbase $l$, and force/torque coefficient of motor denoted as ${K_{\rm{T}}}$ and ${K_{\rm{M}}}$, and $z_{\rm{t}}$ denotes the distance from the force plane to CoB, as shown respectively from (\ref{eq_B1}) to (\ref{eq_B3}). $e_{\rm{x}}$ denotes the unit directional row vector for $x$ direction, so as the other states, where $e_{\rm{z}}$, $e_{\rm{\phi}}$, $e_{\rm{\theta}}$ and $e_{\rm{\psi}}$ are linearly independent, for example, $e_{\rm{z}} = [1 \ 1 \ 1 \ 1]$, $e_{\rm{\phi}} = e_{\rm{y}} = [\text{-}1 \ 1 \ 1 \ \text{-}1]$, $e_{\rm{\theta}} = -e_{\rm{x}} = [\text{-}1 \ 1 \ \text{-}1 \ 1]$, and $e_{\rm{\theta}} = [1 \ 1 \ \text{-}1 \ \text{-}1]$.

Then the kinematic model converted by the dynamic model can be reorganized as
\begin{align}
     \boldsymbol{\ddot \eta}  &=   \dot {\bm{J}} \bm{\nu} +  {\bm{J}}  \dot {\bm{\nu}} 
    = {\bm{B}}_{\rm{e}} \bm{u} +  \bm{f}(\mit{\bm{\eta}} ,\dot {\mit{\bm{\eta}}} ),
\label{eq_ned}
\end{align}
where 
\begin{equation}
    {\bm{B}}_{\rm{e}}={\bm{J}}{\bm{M}^{-1}}{\bm{B}} (\beta),
\end{equation}
\begin{equation}
    {\bm{f}(\bm{\eta} ,\dot {\bm{\eta}} )} =
    (\dot {\bm{J}}-{\bm{J}}{\bm{M}^{-1}}{\textit{\textbf{C}}} -{\bm{J}}{\bm{M}^{-1}}{\bm{D}}){\bm{J}^{-1}}\dot {\mit{\bm{\eta}}}-{\bm{J}}{\bm{M}^{-1}}
    \bm{g} (\mit{\bm{\eta}}).
\label{eq_ned_f}
\end{equation}

\begin{table}[!t]
\centering
\caption{Details of subsystem}
\centering
\begin{tabular}{*{4}{c}}
\toprule
 Index & Name & States & Inputs  \\
\midrule
1 & Longitudinal & ${\mit{\bm{\eta}}}_1 = [x \ z \ \theta]^{\rm{T}}$ &${\varpi _{1}^2}={\varpi _{3}^2} \ne {\varpi _{2}^2} = {\varpi _{4}^2}$  \\
2 & Transverse & ${\mit{\bm{\eta}}}_2 = [y \ z \ \phi]^{\rm{T}}$ & ${\varpi _{2}^2}={\varpi _{3}^2} \ne {\varpi _{1}^2} = {\varpi _{4}^2}$  \\
3 & Heading & ${\mit{\bm{\eta}}}_3 = \psi$ & ${\varpi _{3}^2}={\varpi _{4}^2} \ne {\varpi _{1}^2} = {\varpi _{2}^2}$  \\
4 & Heave & ${\mit{\bm{\eta}}}_4 = z$ & ${\varpi _{1}^2}={\varpi _{2}^2} = {\varpi _{3}^2} = {\varpi _{4}^2}$  \\
\bottomrule
\end{tabular}
\label{table0}
\end{table}

\subsection{Decoupled Model}
Due to its symmetrical features, Mirs-Alioth can be decoupled into several subsystems for easier analysis, as commonly done in complex nonlinear systems as illustrated in \cite{healey1993multivariable}.
Then the subsystem model can be shown as:
\begin{equation}
   {\boldsymbol{\ddot \eta}}_i = {\bm{B}}_{{\rm{e}}i} \bm{u}_i +  \bm{f}({\mit{\bm{\eta}}}_i ,\dot {{\mit{\bm{\eta}}}}_i ) \quad (i=1,...,4).
    \label{eq_long}
\end{equation}

Table \ref{table0} shows that different subsystems with various input constraints are represented depending on the selection of index $i$, i.e., $i = 1 \triangleq $ longitudinal subsystem, etc.

\section{Underwater Motions Analysis \label{analysis}}
In this section, we thoroughly explore and analyze the underwater motion characteristics of Mirs-Alioth. Through this analysis, we aim to derive insights that guide the design of the controller.
Specifically, first, we derive the singularities caused by morphing in the vehicle using the CEM (\ref{eq7}). Next, this serves as a mathematical tool for a deeper analysis of the underwater motion characteristics of the vehicle. Finally, from the results of the analysis, we gain valuable insights for the design of the controller.

\subsection{Singular Thrust Tilt Angle \label{stta}}
Typically, as a result of the additional tilting mechanism, there exists a series of thrust tilt angles that result in certain states being unable to produce a corresponding response. These angles hereafter be referred to as Singular Thrust Tilt Angle (STTA).
We define a collection of all STTAs of our vehicle as $\bm{T}$, containing four elements derived from subsystem (\ref{eq_long}), 
    \begin{equation}
   \bm{T} = \{ \beta^*_{\rm{\phi}},\ \beta^*_{\rm{\theta}},\ \beta^*_{\psi},\  \beta^*_{\rm{{heave}}} \}, 
    \end{equation}
    where 
    \begin{subequations}
    \begin{numcases}{}
    \beta^*_{\rm{\phi}}  = {\rm{arctan}} \left( \footnotesize{\frac{{  {M_{14}}{K_{\rm{T}}} - {K_{\rm{M}}}{M_{22}} + {z_{\rm{t}}}{K_{\rm{T}}}{M_{22}}}}{{l{K_{\rm{T}}}{M_{22}}}}} \right),\label{eq_hover3_1} \\
    \beta^*_{\rm{\theta}}  = {\rm{arctan}} \left( \frac{{ - {M_{15}}{K_{\rm{T}}} + {K_{\rm{M}}}{M_{11}} + {z_{\rm{t}}}{K_{\rm{T}}}{M_{11}}}}{{l{K_{\rm{T}}}{M_{11}}}} \right),\label{eq_hover3_2} \\
    \beta^*_{\psi} = {\rm{arctan}}({K_{\rm{T}}}l/{K_{\rm{M}}}),\label{eq_hover3_3}\\
    \beta^*_{\rm{{heave}}} = 0.\label{eq_hover3_4}
    \end{numcases}
    \label{eq_hover3}
\end{subequations}

From heading and heave subsystem in Table \ref{table0}, when $\beta = \beta^*_{\psi}$ or $\beta = \beta^*_{\rm{{heave}}}$, these subsystems are reformulated as
\begin{equation}
    (m - {Z_{{\rm{\dot w}}}}) \ddot z + {Z_{\rm{w}}} \dot z - (G-B) = 0,
\label{eq30}
\end{equation}
\begin{equation}
    ({I_{{\rm{zz}}}} - {N_{{\rm{\dot r}}}})\ddot \psi + {N_{\rm{r}}}\dot \psi =0.
\label{eq28}
\end{equation}

From (\ref{eq30}) and (\ref{eq28}), the control input $\bm{u}$ in these original subsystems is canceled out, leading to these channels being unrelated to the control inputs, confirming that these values are STTAs.

For the longitudinal subsystem model, the generated expansion control allocation matrix of longitudinal subsystem ${\bm{B}}_{{\rm{e}}1}$ from (\ref{eq_long}) yields

\begin{footnotesize}
\begin{equation}
    {\bm{B}}_{{\rm{e}}1}=\left[ {\begin{array}{*{20}{c}}
    \frac{\sqrt 2}{2}{c_\theta {e_\theta }( - {M_{55}}c_\beta {{{K}}_{\rm{T}}} - {k_2}{M_{51}}) - s_\theta s_\beta {m_{\rm{k}}}{{{K}}_{\rm{T}}}{e_{\rm{z}} }}\\
    - {  s_\theta {e_\theta }( - {M_{55}}c_\beta {{{K}}_{\rm{T}}} - {k_2}{M_{51}}) - c_\theta s_\beta {m_{\rm{k}}}{{{K}}_{\rm{T}}}{e_{\rm{z}} }}\\
    \frac{\sqrt 2}{2}{{e_\theta }({k_2}{M_{11}} - {M_{15}}c_\beta {{{K}}_{\rm{T}}})}
    \end{array}} \right],
    \label{eq31}
\end{equation}
\end{footnotesize}%
where 
$m_{\rm{k}} = ({M_{11}} {M_{55}}- {M_{51}} {M_{15}})/{M_{33}}$.

From the aspect of the rank of the control allocation matrix directly,
a certain channel cannot have a corresponding response for the longitudinal subsystem model if and only if when ${\bm{B}}_{{\rm{e}}1}$ is rank deficient.
Then from the control allocation matrix (\ref{eq31}), its third row can be reorganised as 
\begin{equation}
    c_\beta= \frac{{k_2}{M_{11}}}{{M_{15}} {{{K}}_{\rm{T}}}}.
    \label{condition_surge_mid}
\end{equation}

Substituting (\ref{eq_B2}) into the term (\ref{condition_surge_mid}), it yields the result of STTA (\ref{eq_hover3_2}).
Similarly, we can also derive the STTA of the transverse subsystem, by substituting (\ref{eq_B1}) into the term (\ref{condition_sway_mid}) of the third row from the control allocation matrix ${\bm{B}}_{{\rm{e}}2}$, 
\begin{equation}
    \frac{\sqrt 2}{2}{{e_\phi }({k_1}{M_{22}} - {M_{14}}c_\beta {K_{\rm{T}}})} = 0.
    \label{condition_sway_mid}
\end{equation}

\subsection{Hovering and Descending/Ascending Motion}
Hovering is essential for new quadrotor-like vehicles, particularly underwater. Traditional underwater quadrotors depend on passive forces like gravity and buoyancy, making key design parameters crucial. Moreover, these vehicles are unable dive without angular motion due to a lack of active $z$-direction force.

Based on the heave subsystem model presented in Table~\ref{table0}, to achieve stable hovering, the vehicle must maintain uniform motor rotational speeds to minimize extraneous torque. Additionally, the force generated in the \( z \)-direction should be sufficient to counterbalance the net force resulting from gravity and buoyancy.

Furthermore, it can be concluded that the control inputs that adhere to the constraints outlined in (\ref{eq_hover5}) and (\ref{eq_hover4}) result in a descending/ascending motion,
    \begin{subequations}
    \begin{numcases}{}
        s_\beta  \varpi^2 \ge (G-B) / 4{K_{\rm{T}}},
      \label{eq_hover5}\\
      \left\{
        \begin{aligned}
        \beta &> 0 \triangleq \rm{ascending}, \\
        \beta &< 0 \triangleq \rm{descending}. \\
        \end{aligned}
        \right.
        \label{eq_hover4}
    \end{numcases}
    \end{subequations}

It is worth noting that, the first boundary implies that the thrust force should be dominant. Otherwise, the vehicle is driven by passive force. 

\subsection{Leveling Motions \label{planar motion}}
Compared with the fixed-tilted vehicle both in the air and underwater, the leveling motions are the most distinct movement. From the analysis above, STTA (singularities) can be used to generate the surge/sway motion with invariant pitch/roll and heave. Generally, it should be noted that due to the symmetry of quadrotor-based locomotion, the transverse subsystem exhibits comparable conditions as the longitudinal subsystem. Thus only the latter are discussed in the following discourse.

One notes that the thrust tilt angle satisfying the constraints in (\ref{eq_constraint1_1}) and the control inputs satisfying (\ref{eq_constraint1_2}) with condition (\ref{eq_constraint1_3}) will generate surge motion without considering disturbance,
\begin{subequations}
    \begin{numcases}{}
    \beta = \beta^*_{\rm{\theta}}, \label{eq_constraint1_1}\\ 
    \sum\limits_4^i {{\varpi _i^2} = \frac{{G - B}}{{ s_\beta {K_{\rm{T}}}}}}, \label{eq_constraint1_2}\\
    {{z_{\rm{g}}}} =  -\frac{{{X_{\dot {\rm{q}}}}}}{m} \label{eq_constraint1_3}.
    \end{numcases}
    \label{eq_constraint1}
\end{subequations}

In order to prove the constraints shown above, substituting constraints $\beta = \beta^*_{\rm{\theta}}$ and ${{z_{\rm{g}}}} =  -\frac{{{X_{\dot {\rm{q}}}}}}{m}$ into the longitudinal subsystem model (\ref{eq_long}), we can get:
\begin{equation}
\ddot \theta  =  {k_{z0}}\dot z + {k_{\theta 0}}\dot \theta,
\label{eq_theta_ddot}
\end{equation}
where  ${k_{z0}} =  {C_{18}}{M_{11}} $, and ${k_{\theta 0}} = {M_{15}}{C_{12}} - {M_{\rm{q}}}{M_{11}}$. 

Similarly, substituting constraint (\ref{eq_constraint1_2}) into heave motion of the longitudinal subsystem model when the initial heave velocity is zero, then the absence of heave motion is observed.
Combining what is illustrated above with (\ref{eq_theta_ddot}), there is no pitching motion but only surge motion exists.

Furthermore, some complex motions are generated by compositing them. Theoretical, these motion can be achieved by ensuring that the constraint in (\ref{eq_constraint1_2}) controls the net force in the \( z \)-direction.
And these motions, is characterized by a substantially linear displacement with minimal angular motion. It is also unattainable for fixed-tilted vehicles due to the limitations imposed by the actuation mechanism.

\begin{remark}
    From the above analysis, some insights are drawn as follow:
    \begin{enumerate}
        \item It is evident that \( \beta^*_{\theta} \neq \beta^*_{\phi} \) implies that if the control input is \( \beta \), it is not possible for both the roll and pitch channels to simultaneously operate at STTA. Therefore, a logic switching is required.
        \item  From what mentioned in the last item, for one channel is located at singularity, but the control direction is uncertain for the other channel. An extra tool is required to handle the uncertainty.
        \item Even though, the definitions of STTA are unrelated to the disturbance directly. It is perturbed by them as well. Then, controlling the thrust tilt angle into certain constants are unreliable like \cite{jin2015six}.
        \item When generating composited motions, coupling effects are existed. 
        Even though the control input cannot work on the corresponding channel in $\bm{T}$, the coupling term might perturb the corresponding motion.
        Significantly, these perturbations might deviate the motion out of our expectations. One more controller is required to compensate these effects.
    \end{enumerate}

\end{remark}

\section{Controller Design}\label{controller}

\begin{figure}[h!]
\centering
\includegraphics[width=2.9in]{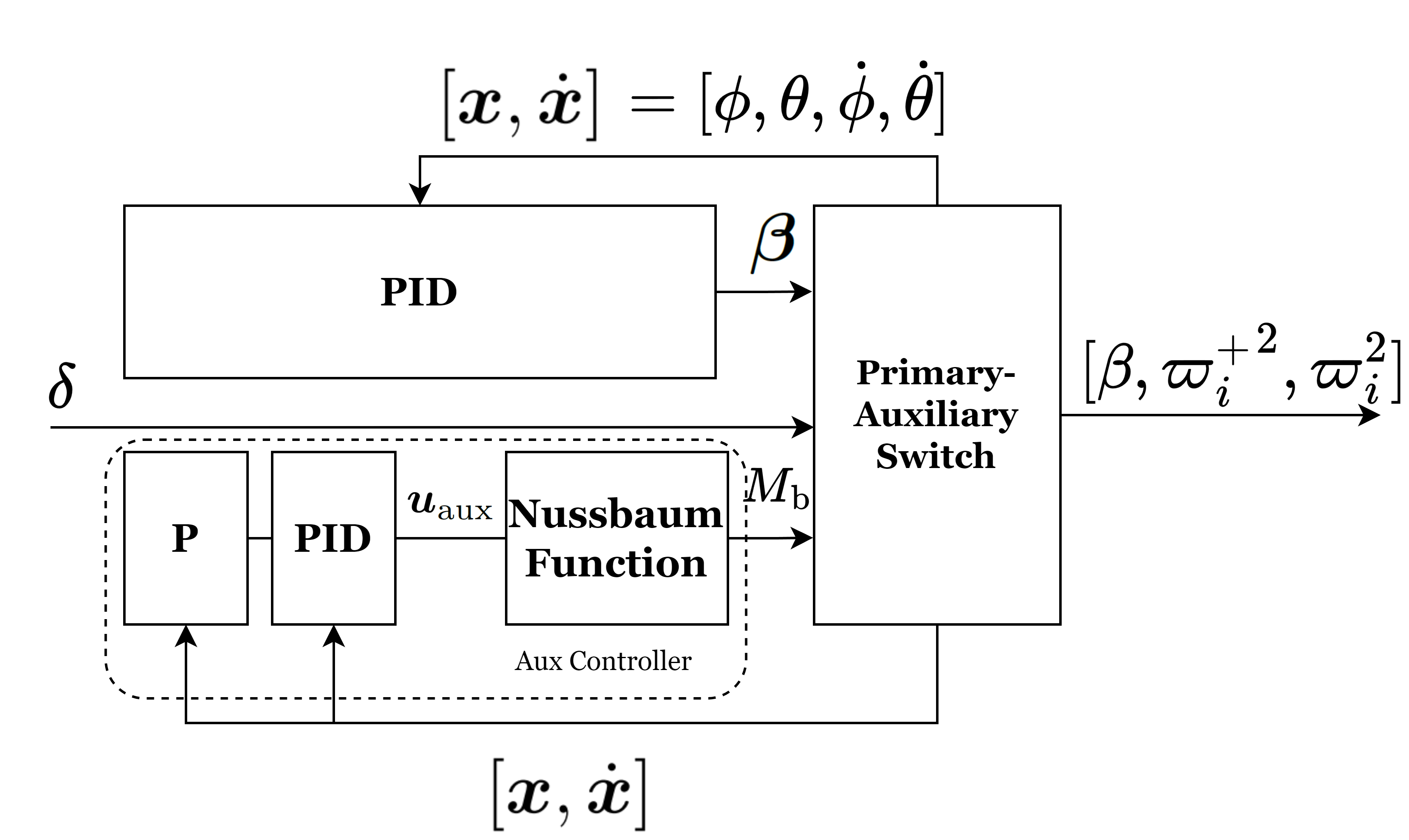}
\caption{Leveling controller diagram.}
\label{fig_control_planar}
\end{figure}

This section introduces the details of the designed controller for operating in the STTA, based on the previous discussion. 
The leveling controller, detailed in Fig.~\ref{fig_control_planar}, regulates the vehicle based on the STTA. Here, some parameters are defined. For example, taking \({\bm{\delta}} = [\delta_{\rm{x}};\delta_{\rm{y}}]\) as desired forces and outputting \(\varpi^+_i\) as motor speed via an auxiliary controller.

As presented in Fig.~\ref{fig_control_planar}, a PID controller is specifically deployed for attitude regulation within the STTA framework by
\begin{equation}
     \bm{\beta} = -{\rm{sgn}}({\bm{\delta}})\left({\textbf{K}}_{\bf{P}\beta}{\bm{x}} +  {\textbf{K}}_{\bf{D}\beta}{\bm {\dot{x}}} + {\textbf{K}}_{\bf{I}\beta}\int{\bm {x}}\rm{d}t\right),
\label{beta_PID}
\end{equation}
where $\bm{\beta} = [\beta_\theta, \beta_\phi]$, \(\textbf{K}_{\bf{P}\beta}\), \(\textbf{K}_{\bf{D}\beta}\), and \(\textbf{K}_{\bf{I}\beta}\) are the proportional, derivative, and integral gains for controller, respectively. ${\rm{sgn}}(\cdot)$ is a sign function.

As stated in Section \ref{stta}, the control outputs in \( \bm{\beta} \) cannot be used to control two separate channels (roll and pitch) simultaneously. Consequently, a primary channel and an auxiliary channel are introduced, which can switch between each other depending on specific operational scenarios, as shown in Algorithm \ref{PA_switch}.
In addition, an auxiliary controller is designed to ensure the stability of the auxiliary channel while the primary channel is regulated by the controller in (\ref{beta_PID}). This is further complicated by the uncertainty in the control direction, which is due to the variable nature of \( \bm{B}_{\text{e}} \), as outlined in (\ref{eq_ned}). To address these issues while maintaining the performance of the higher-level controller, a saturated Nussbaum gain function in \cite{nussbaum1983some,chenci} is employed as 
\begin{equation}
    \bm{M}_{\rm{b}} =  {\bm{\mathcal{N}}}(\zeta_{\phi}, \zeta_{\theta}) \bm{u}_{\rm{aux}},
\end{equation}
\begin{equation}
    {\bm{\mathcal{N}}}(\zeta_{\phi}, \zeta_{\theta}) = {\rm{diag}}[\mathcal{N}(\zeta_{\phi}), \mathcal{N}(\zeta_{\theta})],
\end{equation}
\begin{equation}
    \mathcal{N}(\zeta_{\dagger }) = {\rm{Sat}}( {\rm{cos}}\frac{\pi \cdot \zeta_{\dagger }}{2} \cdot e^{\zeta_{\dagger }^2}),
\end{equation}
\begin{equation}
    \dot \zeta_{\dagger } = {\textbf{K}}_{\zeta} \cdot |{^{\rm{b}} {\bm{\omega}}}_{\dagger }| \cdot |{\bm{u}_{\rm{aux,\dagger }}}|,
\end{equation}
where \( \dagger  \in [\phi, \theta] \) stands for the specific element index in a vector, \( {^{\rm{b}} {\bm{\omega}}}_{\dagger } \) is the current angular velocity, \( \bm{u}_{\rm{aux}} \) is the output of the PID controller, \( \mathcal{N}_{\dagger }(s_{\dagger }) \) stands for the Nussbaum function for different channels, with \( \zeta \) as the state of this function, modulated by a gain matrix \( \textbf{K}_{\zeta} \), and \( \text{Sat}(\cdot) \) is a saturated function.

\begin{algorithm}
\caption{Primary-Auxiliary Switch}
\label{PA_switch}
\begin{algorithmic}[1]
\If{\( |\delta_{\rm{x}}| > |\delta_{\rm{y}}| \)}
    \State \( \beta =  \beta_\theta \quad \text{//Pitch is primary channel}\);
    \State \( \dagger  = \phi \quad \text{//Roll is auxiliary channel}\);
\Else
    \State \( \beta =  \beta_\phi \quad \text{//Roll is primary channel}\);
    \State \( \dagger  = \theta \quad \text{//Pitch is auxiliary channel}\);
\EndIf
\State \( {\varpi^+_i}^2 = \text{Mixer}(\bm{M}_{\rm{b,\dagger }}) \);
\State \( \varpi_i^2 = \text{Mixer}(\bm \delta) + {\varpi^+_i}^2 \);
\end{algorithmic}
\end{algorithm}

Subsequently, a logical state machine termed `Primary-Auxiliary Switch' is necessitated, as outlined in the Algorithm \ref{PA_switch}, to preclude the simultaneous execution of roll and pitch motions. This state machine serves a dual function: it not only ensures the mutual exclusivity of roll and pitch motions but also facilitates the dynamic interchange between the primary and auxiliary control channels, and passed the values to the mixer module in PX4 flight stack eventually \cite{px4}.

\section{Experiment Validations}\label{validation}
\begin{figure}[h]
    \centering
    \includegraphics[width=0.9\linewidth]{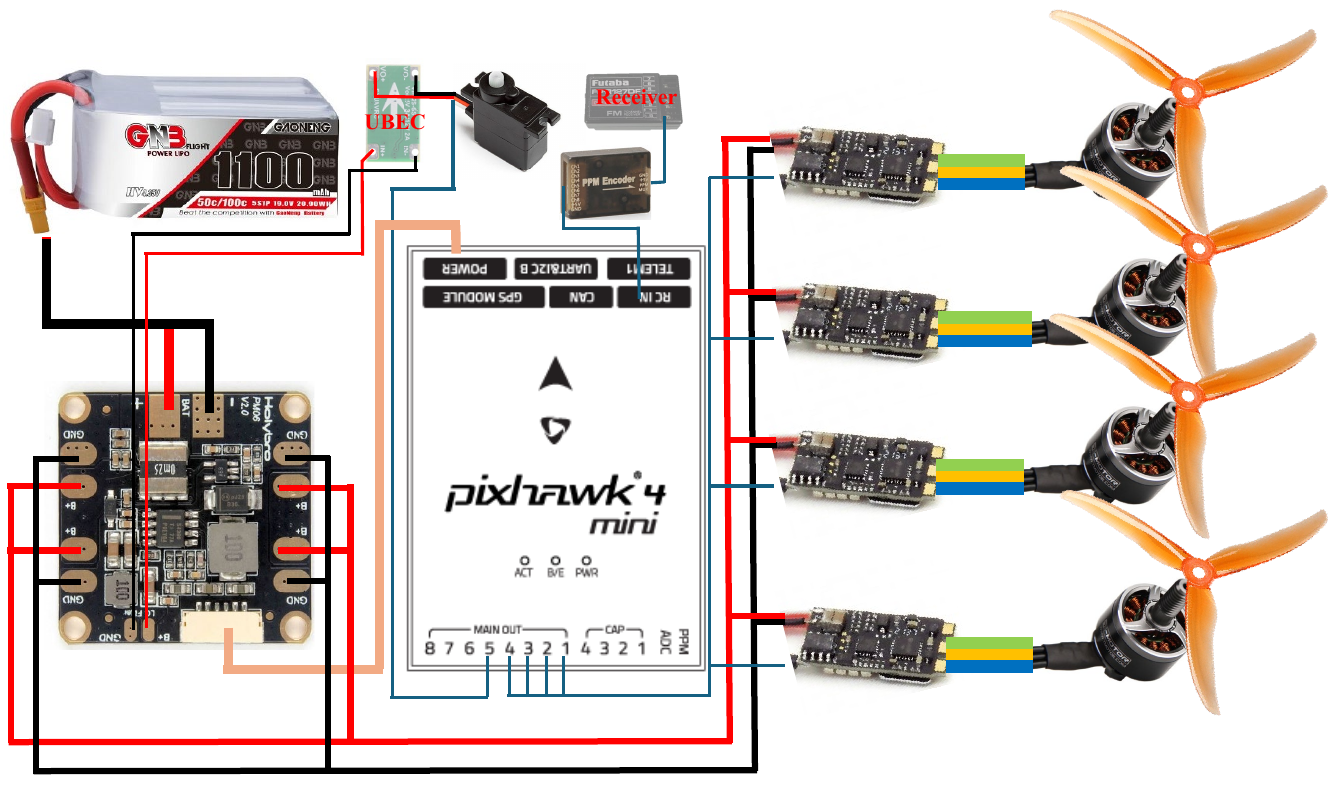}
    \caption{Avionics setup of Mirs-Alioth}
    \label{fig:av}
\end{figure}

In this section, we discuss the experimental setup and results of the proposed controller running on the Mirs-Alioth. The avionics system setup for Mirs-Alioth is shown in Fig.~\ref{fig:av}. The system is powered by a GAONENG 4s 1100mAh battery as the power source, with a Power Distribution Board (PDB) supplying power to each module. The tilt mechanism is powered by a servo motor, with voltage regulated via UBEC to fall within its operational range. The Pixhawk 4 mini serves as the main control board. The propulsion system consists of four HOBBYWING ESCs and T-MOTOR 1507 motors paired with 5146 propellers. Finally, a PPM encoder and receiver, paired with a FUTABA 7CHP, form the signal transmission system, which is designed to work underwater.

Through the aforementioned hardware configuration, we conducted a series of experiments in a water tank. Snapshots and results from the experiments are shown from Fig.~\ref{fig:field} and Fig.~\ref{fig:fieldR} below.

In this set of experiments, under the proposed controller, the vehicle completed the full process of lateral slip, lateral slip turning, transition from lateral to longitudinal slip, longitudinal slip, and longitudinal slip turning. The overall performance can be observed from the snapshots in Fig.~\ref{fig:field}. Specifically, during the first 10 seconds of the experiment, the vehicle is in lateral slip (turning) mode, while for the remaining time, it is in longitudinal slip (turning) mode..

From the roll and pitch angle variations during the experiment, we can see that the attitude angles remained stable without significant fluctuations. For the pitch channel, although there is a relatively large angle amplitude at the beginning, it does not exceed 5 degrees. At around 10 seconds, when the pitch channel becomes the primary channel for turning, there is a brief period of oscillation, but it is quickly stabilized by the controller. On the other hand, the roll channel experience much smaller variations throughout the experiment. This is partly because the moment of inertia for the roll channel is smaller than that of the pitch channel, making it easier for the controller to stabilize.

Furthermore, from the two Nussbaum function curves in Fig.~\ref{fig:fieldR}, we can draw additional insights. For example, the Nussbaum function in the pitch channel adjusts the control direction more frequently than in the roll channel. The pitch channel's Nussbaum function exhibit positive and negative adjustments, while the roll channel does not. This suggests that when the roll channel is the primary channel, the auxiliary channel (pitch) experiences control direction uncertainty. Conversely, when the pitch is the primary channel and the roll is auxiliary, no control direction uncertainty occurs. In this case, the Nussbaum function mainly compensates for uncertainties caused by system nonlinearity. This observation aligns with the motion analysis results stated above.

\begin{figure}[t!]
        \centering
		\subfigure[\label{fig:field1}] 
{\includegraphics[width = 0.24\textwidth]{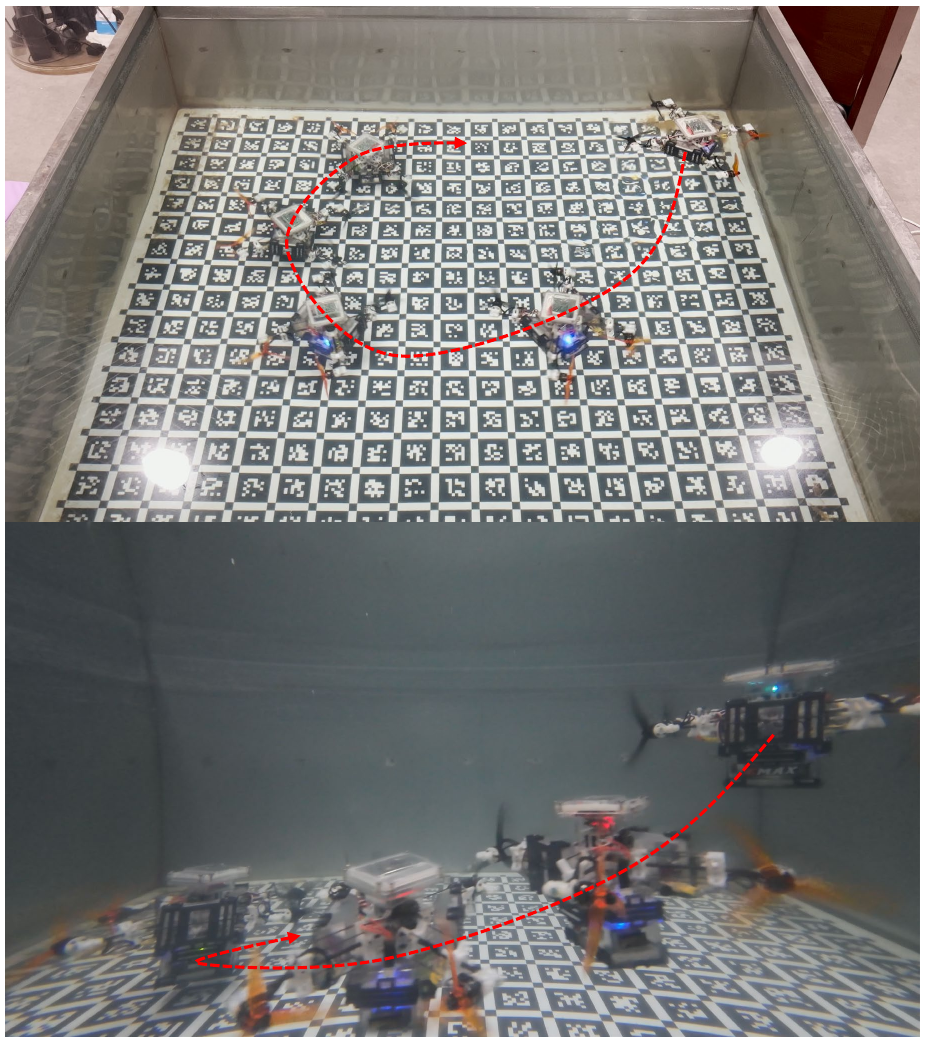}}~
		\subfigure[\label{fig:field2}]
{\includegraphics[width = 0.235\textwidth]{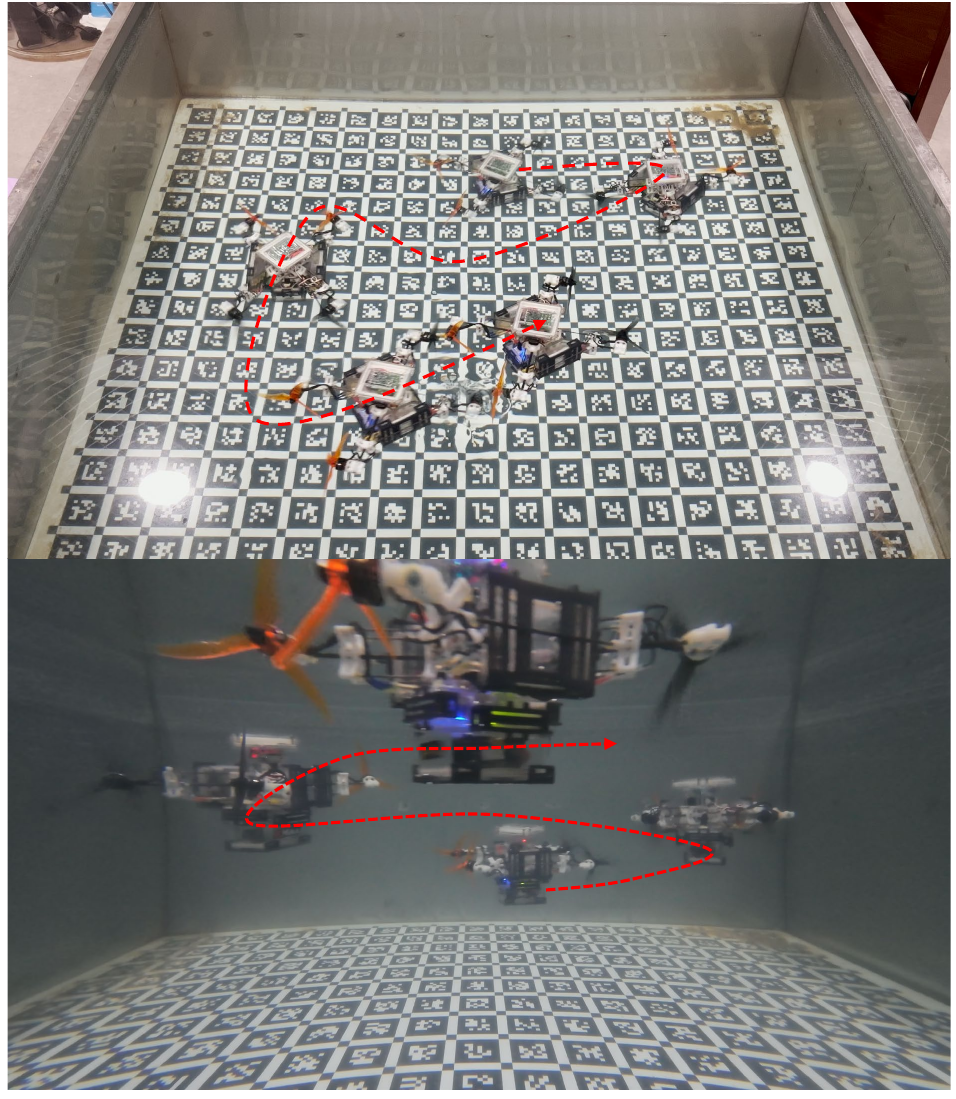}}
\caption{Experiment of both lateral slip (turning) motion and longitudinal slip (turning) motion. (a) Snapshots of lateral slip (turning) motion in top and underwater views. (b) Snapshots of longitudinal slip (turning) motion in top and underwater views. Red dot line denotes the trajectory of the vehicle. }
\label{fig:field}
\end{figure}

\begin{figure}[t!]
    \centering
    \includegraphics[width=0.95\linewidth]{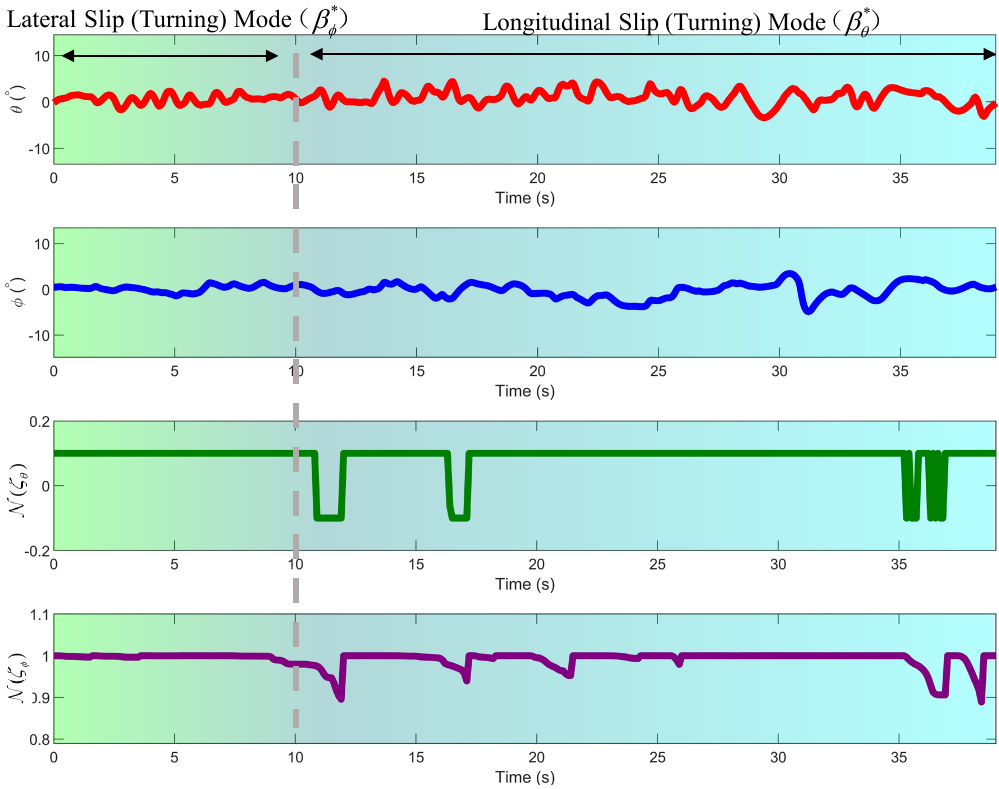}
    \caption{Results of the complete lateral slip (turning) motion and longitudinal slip (turning) motion cycle. Colorful backgrounds correspond to different modes.}
    \label{fig:fieldR}
\end{figure}

\begin{figure}[t!]
    \centering
    \includegraphics[width=0.6\linewidth]{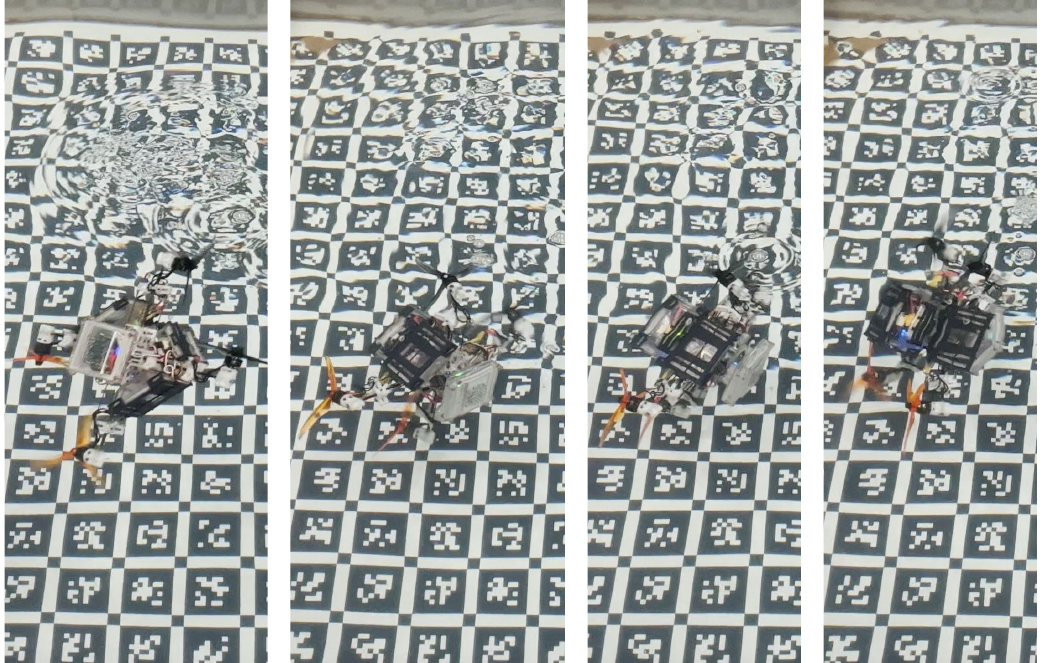}
    \caption{Snapshot of the fail moment in the experiment without the Nussbaum function.}
    \label{fig:fail1}
\end{figure}

\begin{figure}
    \centering
    \includegraphics[width=0.6\linewidth]{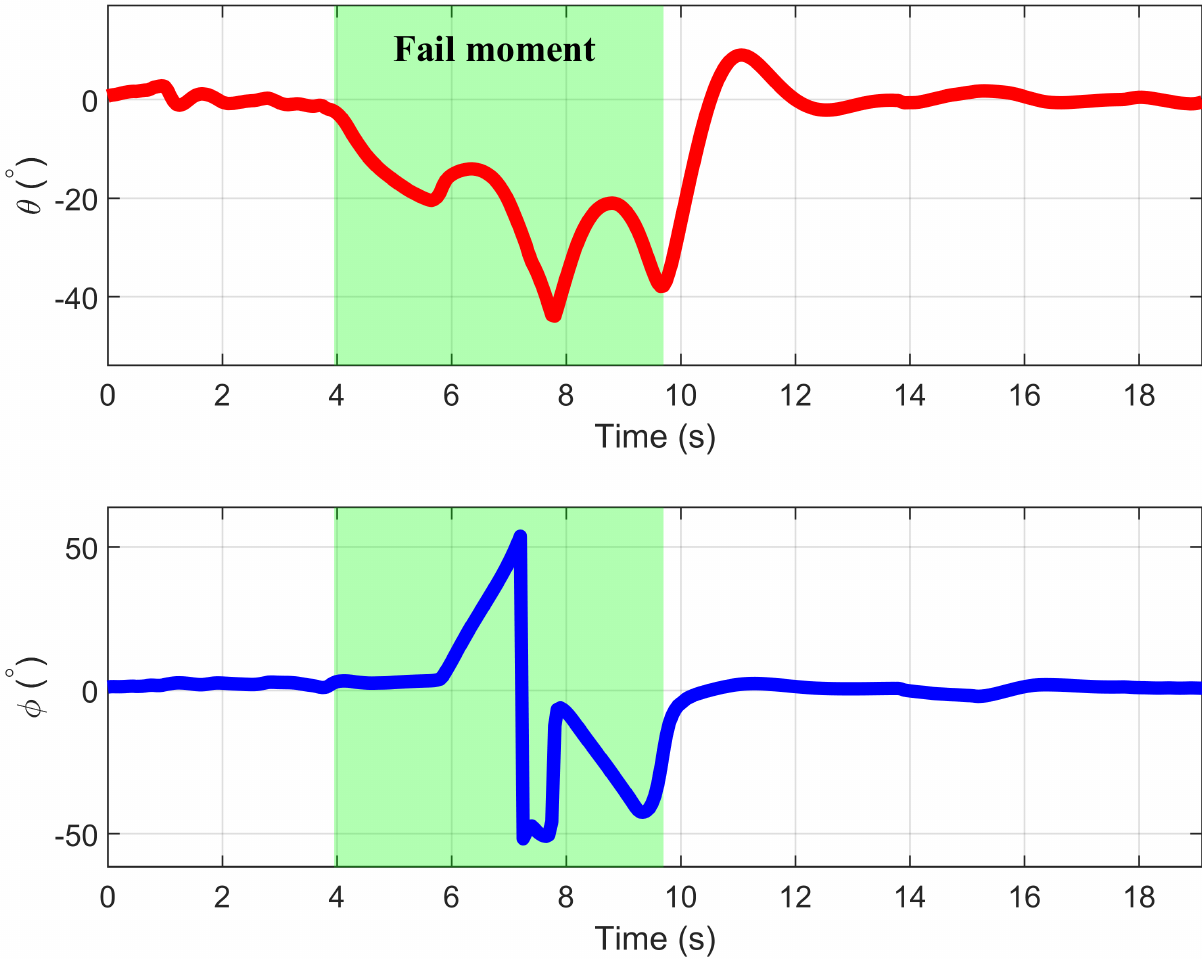}
    \caption{The attitude response during the experiment without Nussbaum function.}
    \label{fig:fail2}
\end{figure}

Building on the previous experiments, we conduct an additional set of ablation experiments. In this set, we compare the performance of the system without the Nussbaum function in Fig.~\ref{fig:fail1} and Fig.~\ref{fig:fail2}.
In this set of ablation experiments, we compare the performance without incorporating the Nussbaum function, analyzing how its absence impacts overall control stability and performance. As shown in the snapshot from Fig.~\ref{fig:fail1}, the performance without the Nussbaum function in the auxiliary controller leads to a catastrophic failure. The curve illustrates that in the green-shaded section, the system becomes unstable, with both channels showing varying degrees of divergence. The input is cut off at 10 seconds, and the system only stabilizes due to its mechanical characteristics. This experiment highlights the necessity of the Nussbaum function in maintaining system stability.

\section{Conclusion}

In this paper, we have conducted a necessary motion analysis of Mirs-Alioth, designed a controller based on this analysis, and validated its effectiveness through experiments. Specifically, we have summarized the motion characteristics of Mirs-Alioth as Singular Thrust Tilt Angles (STTAs) and used this concept as a tool for motion analysis. This analysis revealed four critical factors for generating leveling motion: logic switching between multiple controllers, the use of the Nussbaum function to suppress control direction uncertainty, the need for an additional controller to maintain the thrust tilt angle near singularities, and the management of coupling effects through an auxiliary controller.
Based on these insights, we have designed a controller for Mirs-Alioth and conducted two sets of experiments to verify its effectiveness. The first experiment involved a complete process covering all actions required for leveling motion, with attitude angle fluctuations remaining within 5 degrees throughout. The second experiment was a comparison test without the Nussbaum function. The results demonstrated instability in the vehicle, highlighting the necessity of the Nussbaum function. For future work, we will focus more on real-world robotic testing to lay the foundation for practical applications.

\bibliographystyle{IEEEtran}
\bibliography{mybib}

\end{document}